\documentclass[10pt, twocolumn, letterpaper]{article}

\usepackage{array}
\usepackage{3dv}
\usepackage{times}
\usepackage[utf8]{inputenc}
\usepackage{amsmath}
\usepackage{mleftright}
\usepackage{amssymb}
\usepackage{graphicx}
\usepackage[sort, nocompress]{cite}

\usepackage{xcolor}
\usepackage{eucal}
\usepackage{booktabs}
\usepackage[pagebackref,colorlinks=true,linkcolor=blue,citecolor=blue,urlcolor=blue]{hyperref}
\usepackage{relsize}
\usepackage{microtype}
\usepackage{comment}
\usepackage{multirow}
\usepackage{siunitx}

\threedvfinalcopy %

\def\threedvPaperID{154} %

\ifthreedvfinal\fi

\begin{document}
\frenchspacing

\title{Dimensions of Motion:\\Monocular Prediction through Flow Subspaces}

\author{
Richard Strong Bowen\textsuperscript{*1,2} \quad
Richard Tucker\textsuperscript{*1} \quad
Ramin Zabih\textsuperscript{1,2} \quad
Noah Snavely\textsuperscript{1,2}\\
{\tt\small rsb@cs.cornell.edu \hspace{7em} \{richardt,raminz,snavely\}@google.com}\qquad\qquad\qquad\\[4pt]
\textsuperscript{1} Google Research \quad \textsuperscript{2} Cornell Tech\\
}
\maketitle
\ifthreedvfinal\thispagestyle{empty}\fi
\newcommand{\citefig}[1]{Fig.~\ref{#1}}
\newcommand{\citeeq}[1]{Eq.~\ref{#1}}
\newcommand{\citetab}[1]{Table~\ref{#1}}
\newcommand{\citesec}[1]{Section~\ref{#1}}
\newcommand{\citesecrange}[2]{Sections~\ref{#1}--\ref{#2}}

\newcommand\para[1]{\medskip\noindent\textbf{#1}}
\newcommand\teletype[1]{\texttt{\textbf{\relscale{.98}#1}}}

\newcommand{\mmx}[1]{\mathbf{#1}}
\newcommand{\flow}{\mathrm{\Delta}}
\newcommand{\flowhat}{\hat{\flow}}
\newcommand{\observedflow}{\flow}
\newcommand{\Basis}{\mathcal{B}}
\newcommand{\basis}{\flow} %
\newcommand{\intrinsics}{\mmx{K}}
\newcommand{\pose}{\mmx{P}}
\newcommand{\norm}[1]{\| #1 \|_2}
\newcommand{\mask}{m}

\newcommand{\plocal}{\mmx{M}}
\newcommand{\embedding}{\phi}

\newcommand{\singthresh}{\varepsilon_\text{singular}}

\newcommand\Loss[1]{\mathcal{L}^{\textsf{#1}}}

\def\Image{\mathbf{I}}
\def\Disparity{\hat{\mathbf{D}}}
\def\reals{\mathbb{R}}
\def\Space{\mathcal{S}}
\def\HW2{{H \times W \times 2}}

\newcommand\ablation[1]{\scalebox{.9}[1.0]{\texttt{#1}}}
\newcommand{\tabold}[1]{\textbf{#1}}

\begin{abstract}

We introduce a way to learn to estimate a scene representation from a single image by predicting a low-dimensional subspace of optical flow for each training example, which encompasses the variety of possible camera and object movement. Supervision is provided by a novel loss which measures the distance between this predicted flow subspace and an observed optical flow.
This provides a new approach to learning scene representation tasks, such as monocular depth prediction or instance segmentation, in an unsupervised fashion using in-the-wild input videos without requiring camera poses, intrinsics, or an explicit multi-view stereo step. We evaluate our method in multiple settings, including an indoor depth prediction task where it achieves comparable performance to recent methods trained with more supervision.
Our project page is at \url{https://dimensions-of-motion.github.io/}.
\end{abstract}

\renewcommand{\thefootnote}{\fnsymbol{footnote}}
\footnotetext[1]{Authors contributed equally.}
\renewcommand{\thefootnote}{\arabic{footnote}}

\begin{figure}[h]
\newcommand{\iw}{\dimexpr 0.25\linewidth - .5pt\relax}
\centering
    \begin{tabular}{cc}
         \includegraphics[width=0.35\linewidth]{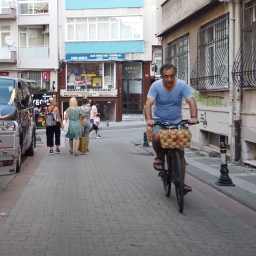} &
         \includegraphics[width=0.55\linewidth]{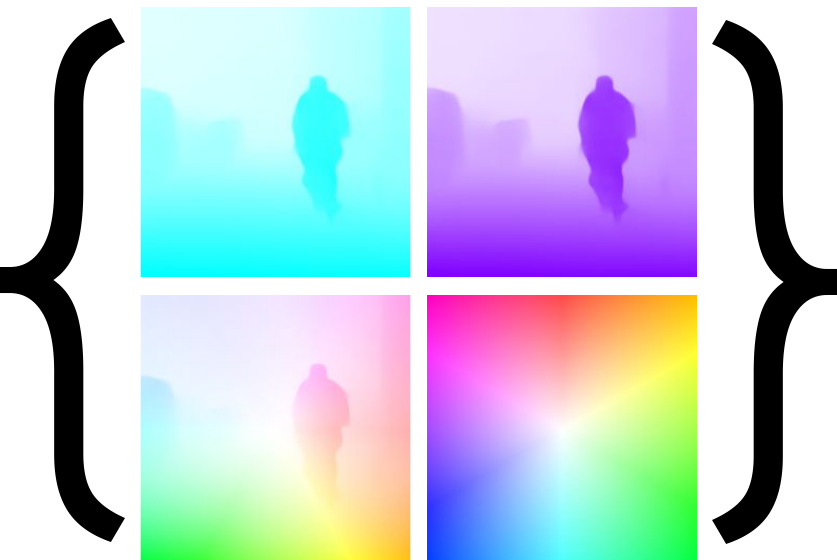} \\
         \scriptsize An example scene & \scriptsize Basis flow fields
    \end{tabular}
    \vspace{.3em}
    \caption{
    This single frame from a video shows a cyclist in a street scene.
    From this image alone,
    can we predict what the optical flow would be between it and the next frame of the video? This single-view flow prediction problem is inherently ill-posed---there could be many possible flow fields, depending on whether the camera is moving left, up, forward, etc, and depending on the motion of the objects in the scene. However, the \emph{space} of possible flow fields should be low-dimensional---that is, it should be spanned by a small number of basis flow fields, some of which (three translations and a rotation) are shown to the right. Moreover, these basis flows will be related to scene properties like depth and instance segmentation, which is why for instance some of these flow fields resemble disparity maps. In this paper, we show how to use these low-dimensional flow subspaces to
    learn to predict properties like depth from single images, supervised by optical flow computed from Internet videos.
    \emph{Video imagery used under Creative Commons license from YouTube channels POPtravel (Figs.~\ref{fig:teaser},~\ref{fig:system},~\ref{fig:walking},~\ref{fig:walking2}) and SonaVisual (\citefig{fig:re10kcc}).}}
    \vspace{-1em}
    \label{fig:teaser}
\end{figure}

\section{Introduction}
Monocular video is widely used as training data in self-supervised learning of depth prediction from single images. Many such methods (including those that predict additional scene properties such as moving object masks) operate by reconstructing one view from another and are supervised using a photometric loss. To perform this reconstruction, either ground truth camera poses and intrinsics are computed in a pre-processing step like structure from motion or SLAM, or else additional networks are trained to predict camera parameters. Either way, an explicit representation of camera pose and intrinsics is part of the training setup.
 
We investigate an alternative approach that uses optical flow---the apparent movement of pixels between two images or frames of video---as supervision, without requiring the poses of those frames. 
Estimating optical flow is still a challenging task, 
but recent deep learning approaches can 
produce quite high-quality two-frame optical flow, and achieve good generalization across datasets. 
How can we use such optical flow from pairs of video frames to help supervise \emph{single}-image tasks like monocular depth prediction?
If the camera is moving between the pair of frames, then the induced flow will be related to the 
scene depth,
and so we might imagine that the problem of \emph{single-image flow prediction} would be a good proxy for other scene prediction tasks. 
However, the task of predicting 
optical flow from a single image
is inherently ill-posed, because an infinite family of possible flows could result from different combinations of camera and scene motion. 
Our approach, then, is not to predict a particular optical flow, nor even a distribution over optical flows, but to predict a low-dimensional flow subspace (a subspace of the much larger space of all theoretically possible optical flows) that contains all realizable instantaneous optical flows (i.e., realizable pixel velocities under small camera or scene motion) given an input image. This overall idea is illustrated in \citefig{fig:teaser}.

In fully unconstrained videos the possible optical flows in a scene are numerous and 
varied, but prior work has shown that under assumptions of instantaneous flow and a rigid scene, the possible flows form a low-dimensional linear subspace, parameterized by depth or disparity. In settings with potentially moving objects, flow resides within a larger but still low-dimensional subspace which we show can be elegantly parameterized by depth and an object embedding.

We use the novel task of predicting a flow subspace as a proxy to learn to predict depth and objects without using any ground truth labels for them, and without requiring camera poses or estimating them via another network. We predict a subspace which encompasses the possible optical flows from any camera movement and focal length, and employ a simple but novel loss that measures the distance between this subspace and the actual optical flow to a nearby frame (computed using a state-of-the-art method such as RAFT~\cite{Teed:RAFT:ECCV20}). This allows us to establish a training setup in which the only required input is video frames.

We review in \citesec{sec:methods} the families of flow that arise from camera movement, and extend this analysis to consider moving objects using an \emph{object instance embedding} (\citesec{sec:object_basis}). We show how this analysis can be applied---in tandem with a linear solver---to train a deep network to predict scene properties from a single image using pairs of frames from Internet videos as supervision, and conduct experiments on depth prediction and object embedding with the RealEstate10K \cite{Zhou:TOG18} dataset and with a varied dataset of videos of people walking around cities. \citefig{fig:system} shows an overview of this training setup. On RealEstate10K we obtain comparable performance to other methods on the same dataset, even without using pose or sparse depth supervision (\citesec{sec:ibims}).

\begin{figure*}[t]
\includegraphics[width=\textwidth]{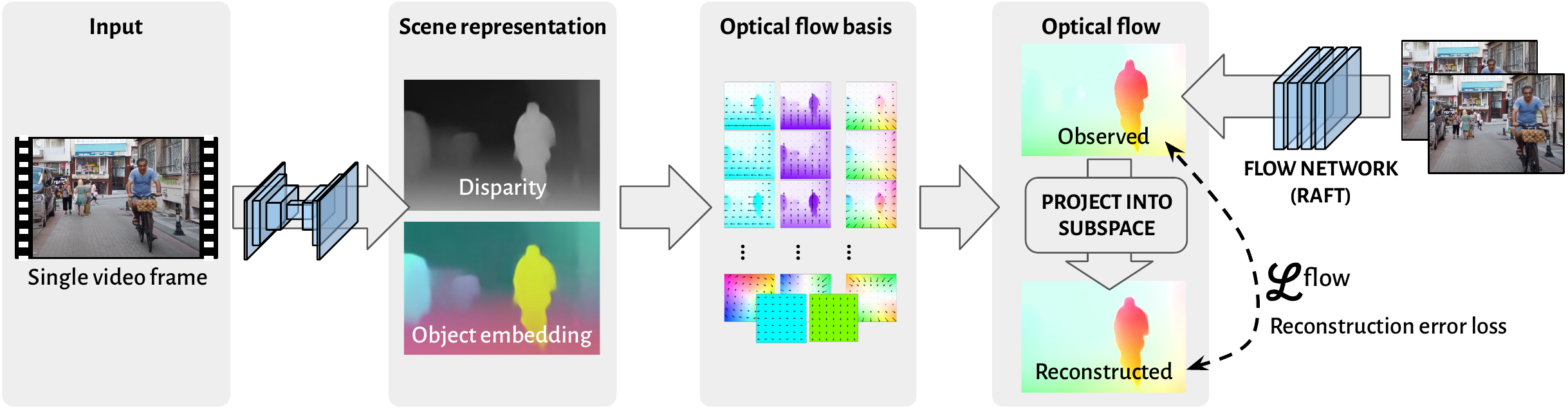}
\caption{System overview. From a single input frame, our network predicts a scene representation consisting of disparity and (optionally) an embedding of objects into ambient space, from which we generate a basis for a subspace of optical flow. During training, we minimise the distance from this subspace to observed optical flow, computed from the input frame and another frame using a pre-existing network.}
\label{fig:system}
\end{figure*}

\section{Related work}
\label{sec:related}
\subsection{Optical flow}
Our method relies on optical flow as a source of supervision; modern two-frame optical flow methods \cite{Teed:CVPR21,Deqing:PWC-Net:CVPR18} are robust and generalize fairly well across datasets.

A number of flow estimation techniques exploit the relationship between optical flow, scene geometry (i.e.~depth) and motion, first analyzed in the context of the human eye~\cite{longuet1980interpretation}. Irani constrains the task of flow estimation between two images using a subspace formulation for instantaneous flow due to camera movement in a rigid 3D scene~\cite{irani2002multi}. The rigidity constraint can be relaxed by treating the subspace as a per-pixel basis for flow and applying regularization to the basis coefficients~\cite{nir2008over}, or by using non-rigid models that allow objects to deform~\cite{garg2010dense}. Flow rank constraint techniques can also be used for tracking and reconstruction~\cite{brand2001morphable,torresani2001tracking,torresani2002space}.

The relationship between optical flow and object or camera movement has also been applied to compute an `ideal' flow for the purpose of comparing and evaluating different flow estimation methods~\cite{mammarella2012comparing}, or to recover the underlying camera and scene parameters from flow~\cite{heeger1992subspace}. More recently, deep learning methods have used this relationship to estimate optical flow simultaneously with object or motion segmentation~\cite{wulff2017optical} and camera movement~\cite{ranjan2019competitive}. Other uses of flow subspaces include building a higher-dimensional flow subspace (dimension 500) by applying PCA to a collection of films to facilitate more efficient flow estimation~\cite{Wulff:PCA-Flow:CVPR15}, and expressing local phenomena such as affine motion and motion edges using a basis of `steerable' flow fields, with the aim of using the decomposition for motion recognition~\cite{FleetBYJ:IJCV00}.

Some work has addressed the problem of predicting optical flow or motion fields from a single image, often by supervising from video~\cite{pintea2014dejavu}, with the aim of producing convincing animations from still images \cite{Holynski:CVPR21}, or as an intermediate step in action recognition \cite{Gao:CVPR18}. Rather than predicting a specific optical flow, Walker \etal~\cite{Walker:ICCV15} predict a probability distribution over a quantized coarse flow. These methods are primarily concerned with object motion and suppose a static camera. In addition to methods that predict motion fields, other work directly predicts future frames from a single image~\cite{Xue:PAMI19}. Again, this work often is primarily concerned with object and not camera motion.

We apply the relationship between optical flow and movement, and a subspace characterization like that of Irani~\cite{irani2002multi}, in a new context: rather than attempting to estimate optical flow, we use it as a source of supervision to learn to predict scene structure from a single input image.

\subsection{Monocular depth}

Supervised learning of monocular depth has a long history; supervision may come from  active sensing with LIDAR \cite{Geiger:KITTI13} or structured light \cite{Silberman:ECCV12}, or from human annotators judging which of a pair of points is closer \cite{Fu:DORN:CVPR18}.
As with 
optical flow, it is difficult and expensive to obtain ground truth to support supervised learning: data may be limited in scale, or spatial density, or both. Instead, depth or disparity may be computed from stereo imagery found online~\cite{Xian:RedWeb:CVPR18}, or from 3D movies as in the MiDaS system~\cite{Ranftl2020}. By applying multi-view stereo~\cite{Schonberger:CVPR16}, depth can also be obtained from collections of in-the-wild photos~\cite{Li:megadepth:CVPR18} or from videos of (artifically) static scenes~\cite{Li:challenge:CVPR19}.

When stereo or multi-view imagery with camera poses is available, one can learn depth without explicit depth supervision. Rather, supervision is provided by a photometric loss (reconstructing one view by warping another according to depth)~\cite{garg2016unsupervised,godard2017unsupervised,kim2020a}.
As view synthesis and depth prediction are related tasks, many single image view synthesis methods naturally produce dense depth as an intermediate step or an auxiliary output. Niklaus \etal~\cite{Niklaus:KenBurns:TOG19} produce a pan-and-zoom effect from a single photo; Tucker and Snavely \cite{Tucker:MPI:CVPR20} learn to generate a multiplane image from which depth may be extracted; Li \etal's~\cite{Li:MINE:ICCV21} MINE combines the properties of this representation with those of Neural Radiance fields. These methods may be supervised by a combination of view synthesis and (when available) sparse depth.

The multi-view processing required to compute accurate camera poses can be computationally expensive and require significant manual tuning. Another approach is to train a second neural network to predict relative pose between two or more input images at training time, in tandem with a first network that predicts monocular depth. To handle video sequences featuring non-static scenes, these methods may also predict an `explainability' map~\cite{Zhou:SfMLearner:CVPR17}, an explicit object motion map~\cite{Li:PMLR21}, a self-discovered object motion map~\cite{bian2019unsupervised}, optical flow for moving regions~\cite{ranjan2019competitive}, or rigid-body object transformations~\cite{Vijayanarasimhan:SfMNet:arXiv17}. Zhao~\etal~\cite{zhao2020towards} learn flow and monocular disparity jointly by sampling from dense correspondences to find a pose, with a photometric loss as well as a scale-invariant depth loss. In contrast, our method uses optical flow rather than image reconstruction for supervision, and does not require or predict explicit camera or object poses.

\subsection{Linear subspaces in computer vision}
Linear subspaces underlie a range of other vision problems in addition to 
optical flow. They apply also to the \emph{appearance manifold}~\cite{Murase:IJCV95}: given an image, other images from slightly different camera viewpoints will lie in a low-dimensional subspace (in particular, 6D, corresponding to the six degrees of freedom of camera motion). Samples of the local appearance manifold can therefore be used for 6DoF camera tracking~\cite{Yang:CVPR07}. In 3D processing, functional maps in the span of a small basis can be used for non-rigid point cloud registration~\cite{huang2022multiway}. There are also classic results in the dimensionality of the space of images of a specific scene under any possible illumination~\cite{Shashua:93,Belhumeur:CVPR96,Garg:ICCV09}, and rank constraints have been studied in the context of motion segmentation when considering multiple rigidly moving objects~\cite{Li:CVPR07}.

\subsection{Object instance embeddings}
To handle moving objects, we produce a per-pixel \emph{object instance embedding} (detailed in Sec.~\ref{sec:object_basis}), where pixels in the same object should map to the same vector, while pixels in different objects as well as the background should map to different vectors.
This embedding is related to the 
task of instance segmentation~\cite{hafiz2020survey}. 
While most instance segmentation approaches employ multi-stage pipelines that include mask proposals, clustering, or other complex techniques, we take an approach similar to Fathi \etal~\cite{fathi2017semantic}, who learn an end-to-end, per-pixel 64-dimensional embedding using ground-truth labels (whereas our formulation is self-supervised). A discretization method such as clustering run on such embeddings may produce reasonable instance segmentations. Newell \etal~\cite{newell2017associative} produce, in a supervised way, a per-class heatmap and a per-class, one-dimensional index (or ``tag'') at each pixel to separate instances.

\section{Methods}\label{sec:methods}

Our approach has two main parts. First, in \citesec{sec:method_learning}, we consider the concept of a basis for optical flow, and show a way to train a system that produces such a basis by using observed flow as supervision and learning to minimize a \emph{flow reconstruction error}. 

Then, in \citesecrange{sec:camera_basis}{sec:object_basis}, we identify subspaces of optical flow corresponding to certain assumptions about the scene. In each case we give a basis for optical flow, identify its dimension, and show how it is parameterized by an appropriate scene representation (disparity, object embedding) which could be predicted by a neural network.

\subsection{Learning a subspace of optical flow}
\label{sec:method_learning}
For an input image $\Image$ of size $H \times W$, the space of all possible optical flow fields is $\reals^{\HW2}$, since flow consists of a separate 2D motion vector for each pixel. But only a tiny fraction of these theoretical optical flows are actually realizable given a specific scene. We represent such possible flows as a low-dimensional subspace $\Space$ of this space.  $\Space$ consists of linear combinations of a set of flows $\basis_i$ that form a flow-basis $\Basis$:
\begin{align}
\Basis &= \{\basis_0, \basis_1, \ldots\, \basis_{n-1}\},\;\basis_i \in \reals^{\HW2} \\
\Space &= \textsc{span} (\Basis)
\end{align}
The individual fields $\basis_i$, and hence the subspace $\Space$, are specific to $\Image$ and not global across all images. While in general the set of plausible flows is not a linear subspace, in the \emph{instantaneous flow} limit the space is closed under linear combination; as long as our time interval is such that rotation and forward motion are small, the instantaneous model is a good approximation of flow \cite{longuet1980interpretation,heeger1992subspace,irani2002multi}.

To learn to predict $\Basis$, we quantify how well the space $\Space$ explains an observed optical flow $\observedflow$, by finding $\flowhat \in \Space$ with minimum distance from $\observedflow$ via projection of $\flow$ into $\Space$. We first find an orthonormal basis for $\Space$ via a (differentiable) singular-value decomposition on the matrix of basis vectors $\begin{bmatrix} \basis_0 \vert \basis_1 \vert \cdots \vert \basis_{n-1} \end{bmatrix},$
where each $\basis_i$ is here viewed as a column vector with $2HW$ elements. The left singular vectors form an orthonormal basis for $\Space$, from which we can compute $\flowhat$. For more details, see
\ifx\arxiv\undefined the supplemental material\else Appendix B\fi.

The distance from $\observedflow$ to $\Space$, or equivalently the error in our reconstructed flow $\flowhat$, is the \emph{flow reconstruction loss}:
\begin{equation}
\Loss{flow} = \norm{\observedflow - \flowhat}
\label{eq:flow_reconstruction_loss}
\end{equation}

Because the SVD routine we use is differentiable, gradient can flow back from this loss through to the basis vectors. In lieu of images with ground truth optical flow we sample pairs of nearby frames from video sequences, using one image from each pair to generate the basis $\Basis$ and running a state-of-the-art flow network~\cite{Teed:RAFT:ECCV20} to produce the observed flow $\flow$ from the pair of images.

In practice, the basis $\Basis$ is not the direct output of our network: instead we output a representation of the scene from which $\Basis$ can be directly computed. In the next sections, we therefore consider flow bases corresponding to specific types of motion.
\subsection{Instantaneous flow from camera motion}
\label{sec:camera_basis}

Optical flow arises from the motion of the camera and of objects in the scene. If the scene is stationary, then all flow comes from camera motion, and we can characterize it explicitly.

We will consider the \emph{instantaneous optical flow} at a point in time. Suppose a world point $(x,y,z)$ projects onto the sensor at pixel $(u,v)$ at time $t=0$. The instantaneous flow is the apparent velocity at this pixel: $\flow = (u', v')$. The instantaneous flow is well-studied within computer vision~\cite{heeger1992subspace,irani2002multi} and in other fields~\cite{longuet1980interpretation}. For a given disparity map and camera intrinsics, the instantaneous flow depends \emph{linearly} on the six parameters of translational and rotational velocity, i.e.~the six degrees of freedom in the camera pose. Consequently, the set of possible instantaneous flows forms a linear space with six basis vectors, which we now enumerate. (For more details and derivations, see for example Heeger and Jepsen \S 3~\cite{heeger1992subspace} or Irani, Appendix A~\cite{irani2002multi}.)

\para{Camera translation.}
For translation along each axis ($\textbf{T}x$, $\textbf{T}y$, $\textbf{T}z$), the basis vectors are:
\begin{equation}
\flow_{\textbf{T}x} = \begin{bmatrix} d \cdot f_x \\ 0 \end{bmatrix},
\flow_{\textbf{T}y} = \begin{bmatrix} 0 \\ d \cdot f_y \end{bmatrix},
\flow_{\textbf{T}z} = \begin{bmatrix} d \cdot (c_x - u) \\ d \cdot (c_y - v) \end{bmatrix}
\label{eq:flow_translation} \\[8pt]
\end{equation}
Here $f_x$ and $f_y$ are the x and y focal lengths of the camera, $(c_x, c_y)$ is the principal point, and $d$ (a function of $(u,v)$) is the disparity or inverse depth at $(u,v)$. The translational flow fields are horizontal and vertical for translation in $x$ and $y$, and radial (centered on the principal point) for translation in $z$, and in all three cases the flow is proportional to disparity $d$ since the further away objects are, the less they appear to move when the camera translates. Note that there is the usual scale ambiguity between translation velocity and disparity.

\para{Camera rotation.}
The basis vectors for rotation about the $x$, $y$ and $z$ axes are:
\begin{align}
\flow_{\textbf{R}x} &= \begin{bmatrix} \frac{1}{f_y}(u - c_x)\,(v - c_y) \\
  f_y + \frac{1}{f_y}(v - c_y)^2 \end{bmatrix} \\
\flow_{\textbf{R}y} &= \begin{bmatrix} f_x + \frac{1}{f_x}(u - c_x)^2 \\
  \frac{1}{f_x}(u - c_x)\,(v - c_y) \end{bmatrix} \\
\flow_{\textbf{R}z} &= \begin{bmatrix} \frac{f_x}{f_y}(v - c_y) \\
   \frac{f_y}{f_x}(c_x - u)\end{bmatrix}
\label{eq:flow_rotation}
\end{align}
As expected, flow from rotation does not depend on disparity, since motion induced by pure camera rotation is independent of depth. At large focal lengths, flow from rotation about the $x$ (or $y$) axis is almost vertical (or horizontal) and uniform; at smaller focal lengths the effects of curvature are more apparent, especially at the corners. For rotation around the $z$-axis, flow is circular (or elliptical if $f_x \ne f_y$) around the optical center.

\begin{figure*}[t]
\begin{center}
\centering
\includegraphics[width=\linewidth,clip]{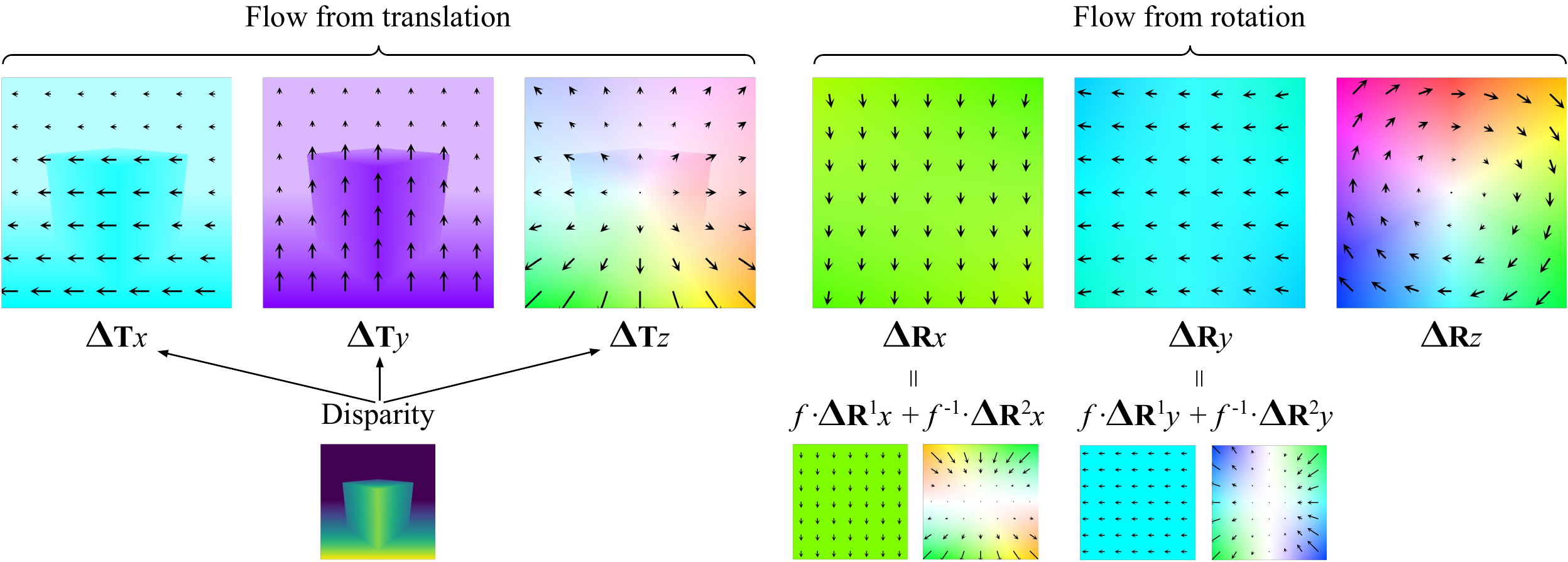}
\caption{Flow basis for camera motion with six degrees of freedom, shown on a simple cube scene (best viewed in color). The translation flows are derived from the depicted scene disparity. See \citesec{sec:camera_basis}.
}
\label{fig:camera_basis}
\end{center}
\end{figure*}

\para{Bases.} Combining translation and rotation, we have a basis for the six-dimensional space of flow due to camera motion:
\begin{equation}
\Basis_\textrm{camera} = \{ \flow_{\textbf{T}x},\; \flow_{\textbf{T}y},\; \flow_{\textbf{T}z},\; \flow_{\textbf{R}x},\; \flow_{\textbf{R}y},\; \flow_{\textbf{R}z} \}.
\end{equation}
\citefig{fig:camera_basis} depicts this flow basis for an example scene.

A common case when dealing with real-world imagery is that $f_x$ and $f_y$ are unknown but equal, and $c_x$ and $c_y$ are known (or assumed to be at the center of the image). Can we produce a basis for flow due to camera motion in this case?
Since basis vectors may be freely scaled up or down, only the flows from rotation about the $x$- and $y$-axes are problematic. We can separate out the terms in $f$ and $1 / f$, replacing each of these two flow fields by a pair (as also shown in \citefig{fig:camera_basis}):
\begin{equation}
\flow_{\textbf{R}x} = f_y\flow_{\textbf{R}^1x} + \frac{1}{f_y}\flow_{\textbf{R}^2x},
\flow_{\textbf{R}y} = f_x\flow_{\textbf{R}^1y} + \frac{1}{f_x}\flow_{\textbf{R}^2y}
\end{equation}
where
\begin{align}
\flow_{\textbf{R}^1x} &= \begin{bmatrix}0 \\ 1\end{bmatrix},\quad 
\flow_{\textbf{R}^2x} = \begin{bmatrix}(u - c_x)\,(v - c_y) \\ (v - c_y)^2\end{bmatrix}, \\
\flow_{\textbf{R}^1y} &= \begin{bmatrix}1 \\ 0\end{bmatrix},\quad
\flow_{\textbf{R}^2y} = \begin{bmatrix}(u - c_x)^2 \\ (u - c_x)\,(v - c_y)\end{bmatrix}.
\end{align}
We thus end up with a basis of eight flow fields, parameterized by disparity $d$, to cover the space of camera movement with unknown focal length. Note that 
this basis actually covers a slightly larger space, since although we assume $f_x = f_y$ we do not have a way to enforce this (nonlinear) constraint in our decomposition of the rotation flows.

\subsection{Instantaneous flow from object motion}
\label{sec:object_basis}
Suppose now that the camera is stationary but that a rigid object in the scene is moving. What does the resulting space of possible optical flow fields look like? For all points $(u, v)$ outside the moving object the flow will be zero. For points \emph{within} the object, we observe that for any rigid object motion (rotation or translation) there is an equivalent camera motion, and thus the space of flow from rigid object motion is exactly the same as the space of flow from camera motion restricted to points in the object. That is, given a binary object mask $\mask$ which is 1 within the object and 0 elsewhere, a basis for rigid movement of the object is given by
\begin{equation}
\Basis_m = \{ \mask\flow \mid \flow \in \Basis_\textrm{camera} \}.
\label{eq:masks}
\end{equation}
Alternatively we may consider a flow basis for \emph{object translation} only, which is just three dimensions per object:
\begin{equation}
\Basis_m^\textrm{translation} = \{\mask\flow_{\textbf{T}x},\; \mask\flow_{\textbf{T}y},\; \mask\flow_{\textbf{T}z}\}.
\end{equation}

\begin{table*}[t]
\centering
\renewcommand{\tabcolsep}{2pt}
\newcolumntype{F}{>{\centering}p{2.8em}}
\resizebox{\textwidth}{!}{\begin{tabular}{@{}lFFFFFFp{.5em}FFFFFF@{}}
\toprule
& \multicolumn{6}{c}{iBims-1} && \multicolumn{6}{c}{NYU Depth V2} \\
\cmidrule{2-7} \cmidrule{9-14}
\hspace{1em}Method (Dataset) & rel$\,\downarrow$ & log10$\,\downarrow$ & RMS$\,\downarrow$ &
$\sigma_1\uparrow$ & $\sigma_2\uparrow$ & $\sigma_3\uparrow$
&& rel$\,\downarrow$ & log10$\,\downarrow$ & RMS$\,\downarrow$ &
$\sigma_1\uparrow$ & $\sigma_2\uparrow$ & $\sigma_3\uparrow$
\tabularnewline
\midrule
\textit{Supervised by depth}\tabularnewline
\hspace{1em}DIW (DIW) \cite{chen2016single} &
   0.25 & 0.10 & 1.00 & 0.61 & 0.86 & 0.95
   && 0.25 & 0.1 & 0.76 & 0.62 & 0.88 & 0.96
   \tabularnewline 
\hspace{1em}MegaDepth (Mega/DIW) \cite{Li:megadepth:CVPR18} &
   0.20 & 0.08 & 0.78 & 0.70 & 0.91 & 0.97
   && 0.21 & 0.08 & 0.65 & 0.68 & 0.91 & 0.97
   \tabularnewline 
\hspace{1em}MiDaS v2.1 (MiDaS 10) \cite{Ranftl2020} &
   0.14 & 0.06 & 0.57 & 0.84 & \tabold{0.97} & \tabold{0.99}
   && 0.16 & 0.06 & 0.50 & 0.80 & 0.95 & 0.99
   \tabularnewline 
\hspace{1em}3DKenBurns (Mega/NYU/3DKB) \cite{Niklaus:KenBurns:TOG19} &
   \tabold{0.10} & \tabold{0.04} & \tabold{0.47} & \tabold{0.90} & \tabold{0.97} & \tabold{0.99}\vspace{4pt}
   && \tabold{0.08} & \tabold{0.03} & \tabold{0.30} & \tabold{0.94} & \tabold{0.99} & \tabold{1.00}
   \tabularnewline 
\midrule
\makebox[0pt][l]{\textit{Supervised by view synthesis plus sparse depth, using pose from SfM}}\tabularnewline
\hspace{1em}Single-view MPI (RE10K) \cite{Tucker:MPI:CVPR20} &
   0.21 & 0.08 & 0.85 & 0.70 & 0.91 & 0.97
   && 0.15 & 0.06 & 0.49 & 0.81 & 0.96 & 0.99
   \tabularnewline 
\hspace{1em}MINE (RE10K) \cite{Li:MINE:ICCV21} &
   \tabold{0.11} & \tabold{0.05} & \tabold{0.53} & 0.87 & \tabold{0.97} & \tabold{0.99}\vspace{4pt}
   && \tabold{0.11} & \tabold{0.05} & \tabold{0.40} & \tabold{0.88} & \tabold{0.98} & \tabold{0.99}
   \tabularnewline 
\textit{Supervised by flow reconstruction only}\tabularnewline
\hspace{1em}Ours (RE10K) &
   0.12 & \tabold{0.05} & 0.55 & 0.85 & \tabold{0.97} & \tabold{0.99}
   && 0.12 & \tabold{0.05} & 0.43 & 0.86 & 0.97 & \tabold{0.99}
   \tabularnewline 
\bottomrule
\vspace{-.5em}
\end{tabular}}
\caption{Depth prediction quality measured with the iBims-1 \cite{koch:2018:ibims} and NYU-V2 \cite{Silberman:ECCV12} benchmarks.
Our method, trained on RealEstate10K, achieves comparable performance with the best methods among those without extra, hard-to-scale supervision (such as structured light, as in NYU). Additionally, we remove the need for an explicit posing or a structure from motion step in the pipeline in, e.g., MINE. See~\citesec{sec:ibims}.}
\vspace{-.0em}
\label{tab:ibims}
\end{table*}

Hence, one way to produce a basis for flow due to object motion would be to predict disparity $d$ and a set of object masks.
But we can instead model potential movers
in the scene \emph{without} 
explicit masks, by introducing an \emph{object instance embedding} $\embedding(u,v) \in \reals^A$. This embedding, like much higher-dimensional embeddings used in instance segmentation~\cite{fathi2017semantic}, gives for each pixel a unit vector in an embedding space of dimension $A$. (In our experiments, $A=6$.) The idea is that pixels in the same object should map to the same point in this space, but that different objects, and the background, should map to different and linearly-independent points.

With up to $A$ objects (including the background) at linearly-independent positions in this space, a matrix $\plocal \in \reals^{3 \times A}$ is sufficient to describe a mapping from embedding space to movement in the $x$-, $y$- and $z$-axes, allowing each object to move independently. The movement at each pixel is then $\plocal\embedding$, and the flow $\flow_{\plocal\embedding}$ due to object and camera translation is:
\begin{align}
\flow_{\plocal\embedding} &= (\plocal \embedding) \cdot \begin{bmatrix} \flow_{\textbf{T}x} \\ \flow_{\textbf{T}y} \\ \flow_{\textbf{T}z} \\ \end{bmatrix} \\
&= \sum_{i=0}^{A-1} \plocal_{0i}\,\phi_i\,\flow_{\textbf{T}x} + \plocal_{1i}\,\phi_i\,\flow_{\textbf{T}y} + \plocal_{2i}\,\phi_i\,\flow_{\textbf{T}z}. \nonumber
\end{align}
Thus a basis for the space of possible flows $\flow_{\plocal\embedding}$ is given by
\begin{equation}
\hspace*{-.5mm} 
\Basis_\embedding^\textrm{translation} = \{ \embedding_i \flow_{\textbf{T}x},\; \embedding_i \flow_{\textbf{T}y},\; \embedding_i \flow_{\textbf{T}z} \mid 0 \leq i < A \}.
\end{equation}
Projecting into this basis implicitly finds a matrix $\plocal$. To allow for camera rotation too we add the various $(\flow_{\textbf{R}\cdot})$ to our basis, giving a basis parameterized by $(d, \embedding)$ with $3A + 3$ dimensions, or $3A + 5$ for unknown focal length. Allowing for \emph{object} rotation is also straightforward (add the $(\embedding_i \flow_{\textbf{R}\cdot})$ to the basis), but we did not find that it improved performance. %

\section{Experiments}

We use the setup described in \citesec{sec:method_learning} to learn monocular depth prediction (\citesec{sec:ibims}) and to learn depth together with object embedding (\citesec{sec:embed}). In either case, our architecture roughly follows DispNet~\cite{mayer2016large}, using an encoder-decoder architecture with skip connections, described more fully in
\ifx\arxiv\undefined the supplementary material\else Appendix A\fi.
For depth, the output is a $H\times W\times 1$ tensor of disparities, with sigmoid activation applied, from which we generate an 8-dimensional basis as described in \citesec{sec:camera_basis}. For instance embedding experiments the output is $H\times W\times (1+A)$, i.e., $1$ dimension for disparity (again with sigmoid activation) and $A$ dimensions for the embedding (normalized to be unit-length at each pixel), from which we generate a $3A + 5$ dimensional basis as described in \citesec{sec:object_basis}. Training data consists of frames from monocular video, with observed optical flow estimated using RAFT~\cite{Teed:RAFT:ECCV20}. No use was made of camera intrinsics, poses, or ground truth depth data in training.

Our implementation is in TensorFlow~\cite{tensorflow2015-whitepaper} and uses \teletype{tf.linalg.svd} for the singular value decomposition (see \citesec{sec:method_learning}).  In our experiments with object embedding, we run the solver twice: once using only the 8-dimensional camera-movement flow basis, and again using the full $(3A + 5)$-dimensional basis, and we train using reconstruction losses from both solves.

\subsection{Learning disparity on static scenes}
\label{sec:ibims}
\begin{figure*}
\centering
\renewcommand{\tabcolsep}{0.5pt}
\renewcommand{\arraystretch}{0}
\newcommand{\iw}{\dimexpr 0.2\linewidth - .5pt\relax}
\small
    \begin{tabular}{ccccc}
         \includegraphics[width=\iw]{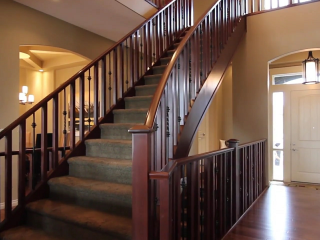} &
         \includegraphics[width=\iw]{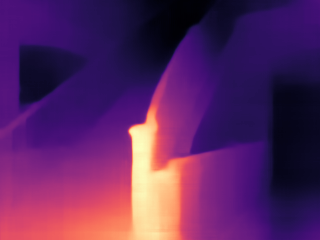} &
         \includegraphics[width=\iw]{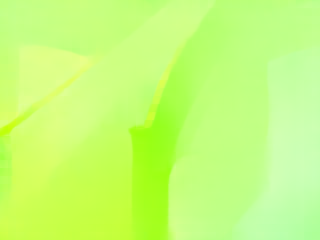} &
         \includegraphics[width=\iw]{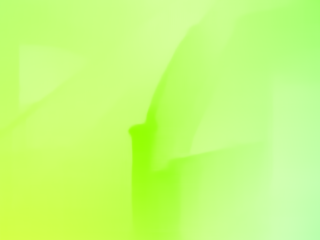} &
         \includegraphics[width=\iw]{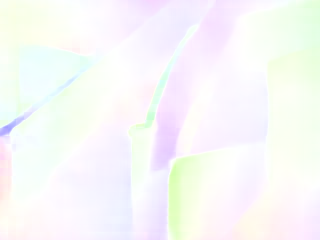} \\ [1pt]
         \includegraphics[width=\iw]{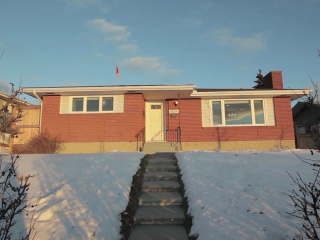} &
         \includegraphics[width=\iw]{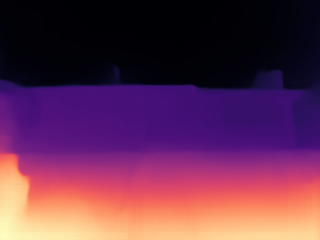} &
         \includegraphics[width=\iw]{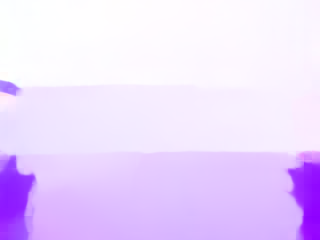} &
         \includegraphics[width=\iw]{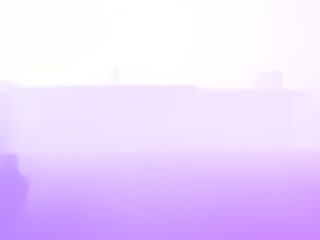} &
         \includegraphics[width=\iw]{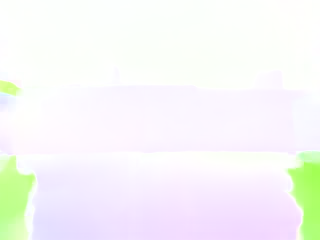} \\ [1pt]
         \includegraphics[width=\iw]{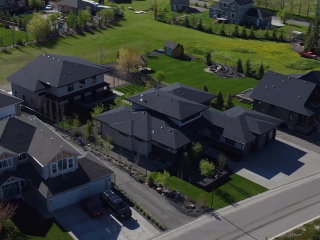} &
         \includegraphics[width=\iw]{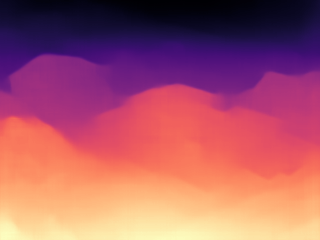} &
         \includegraphics[width=\iw]{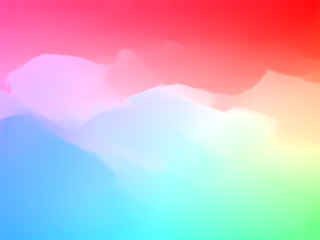} &
         \includegraphics[width=\iw]{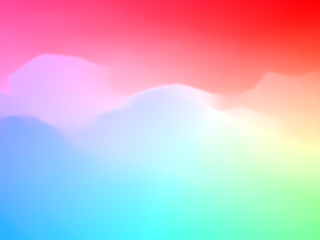} &
         \includegraphics[width=\iw]{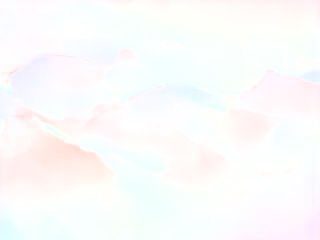} \\ [1pt]
         \includegraphics[width=\iw]{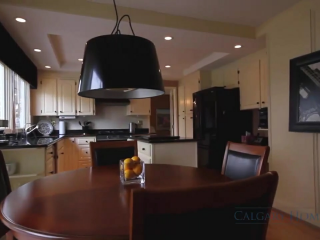} &
         \includegraphics[width=\iw]{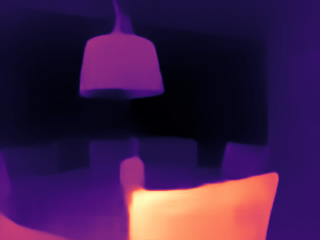} &
         \includegraphics[width=\iw]{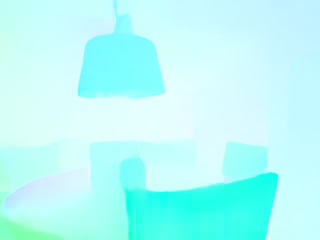} &
         \includegraphics[width=\iw]{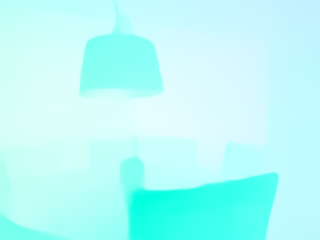} &
         \includegraphics[width=\iw]{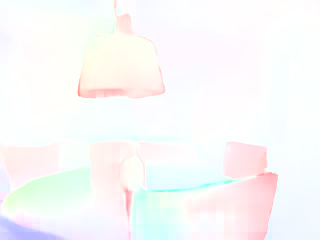} \\ [1pt]
         \includegraphics[width=\iw]{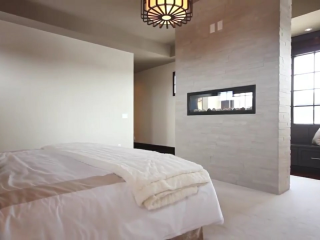} &
         \includegraphics[width=\iw]{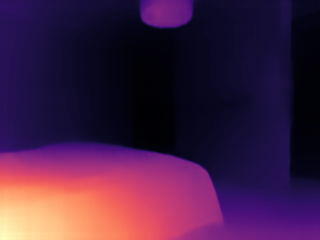} &
         \includegraphics[width=\iw]{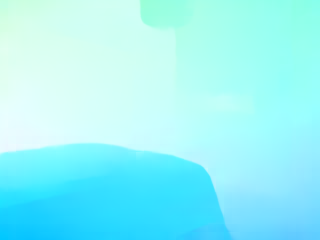} &
         \includegraphics[width=\iw]{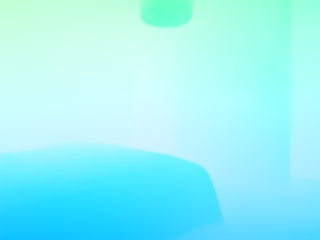} &
         \includegraphics[width=\iw]{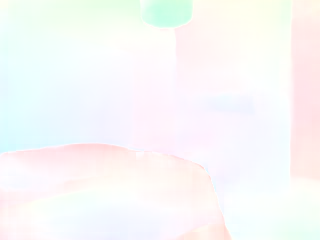} \\ [1pt]
         \includegraphics[width=\iw]{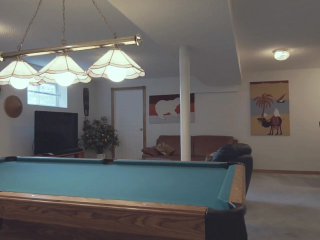} &
         \includegraphics[width=\iw]{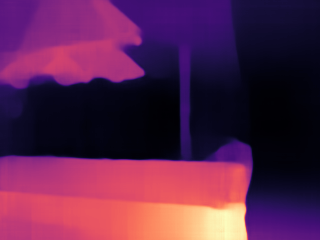} &
         \includegraphics[width=\iw]{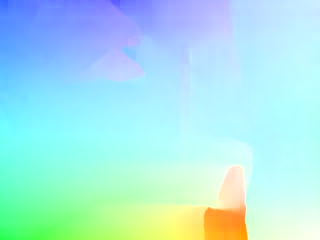} &
         \includegraphics[width=\iw]{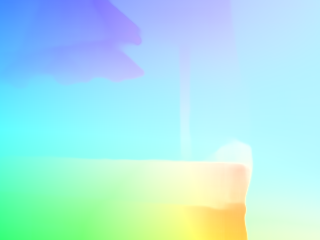} &
         \includegraphics[width=\iw]{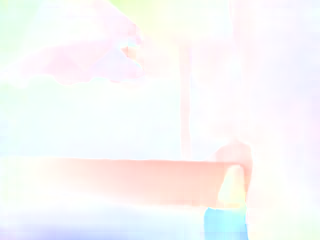} \\ [1pt]
         \includegraphics[width=\iw]{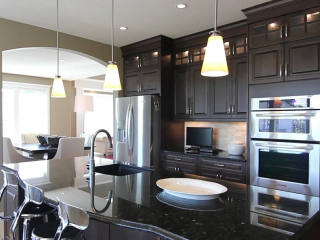} &
         \includegraphics[width=\iw]{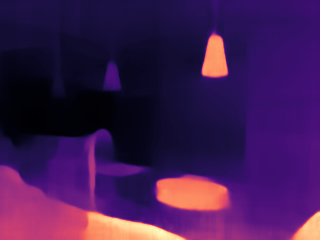} &
         \includegraphics[width=\iw]{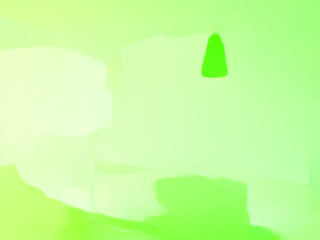} &
         \includegraphics[width=\iw]{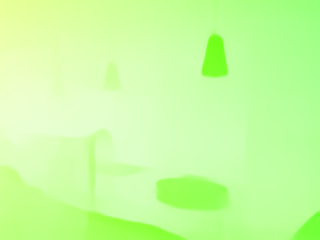} &
         \includegraphics[width=\iw]{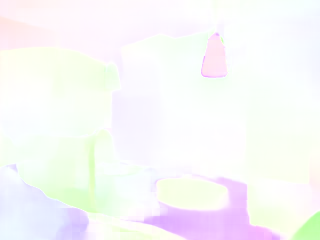} \\ [1pt]
         \includegraphics[width=\iw]{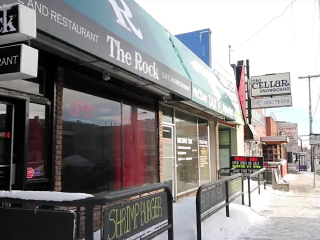} &
         \includegraphics[width=\iw]{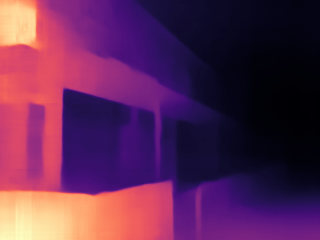} &
         \includegraphics[width=\iw]{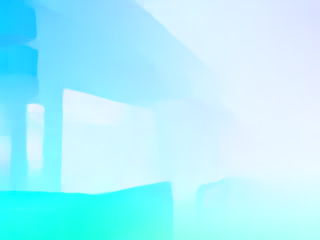} &
         \includegraphics[width=\iw]{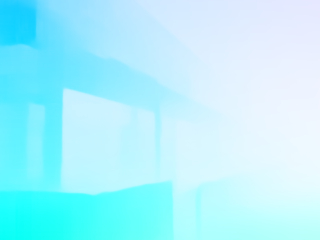} &
         \includegraphics[width=\iw]{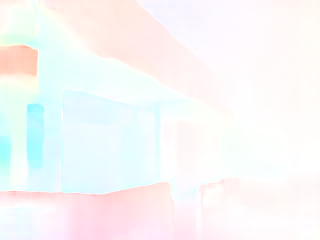} \\ [8pt]
         Input image & Predicted & Flow & Reconstructed & Flow error \\ [5pt]
         & disparity & & flow & \\ [10pt]
    \end{tabular}
    \caption{Experiment on RealEstate10K. Image, disparity predicted by our network, flow, reconstructed flow, and flow error are shown here on test data. (Flow is between source and a target image; the latter is not shown here). See \citesec{sec:ibims}.}
    \label{fig:re10kcc}
\end{figure*}
To investigate depth prediction on static scenes, we train on images from the RealEstate10K dataset~\cite{Zhou:TOG18}, but without making any use of the included camera intrinsics or poses. This dataset consists of frames from internet real estate videos, including both indoor and outdoor settings. Scenes are mostly static and feature a variety of different camera movements. We train on pairs of images 3--10 frames apart, at a resolution of 240$\times$320. We evaluate on the iBims-1~\cite{koch:2018:ibims} and NYU-V2 Depth~\cite{Silberman:ECCV12} test sets, which contain ground truth depth maps, following the protocol of Niklaus \etal~\cite{Niklaus:KenBurns:TOG19}. As shown in \citetab{tab:ibims}, our method achieves comparable performance to other methods trained on this dataset that \emph{do} use camera information or a potentially expensive multi-view stereo or structure-from-motion step in the pipeline. Example outputs are shown in \citefig{fig:re10kcc}.

\subsection{Learning on dynamic scenes}
\label{sec:embed}
\begin{figure}[t]
\centering
\renewcommand{\tabcolsep}{0.5pt}
\renewcommand{\arraystretch}{0}
\newcommand{\iw}{\dimexpr 0.25\linewidth - .5pt\relax}
\small
    \begin{tabular}{cccc}
         \includegraphics[width=\iw]{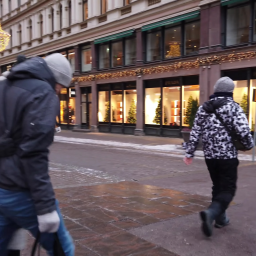} &
         \includegraphics[width=\iw]{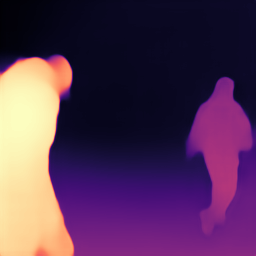} &
         \includegraphics[width=\iw]{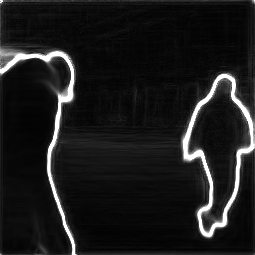} &
         \includegraphics[width=\iw]{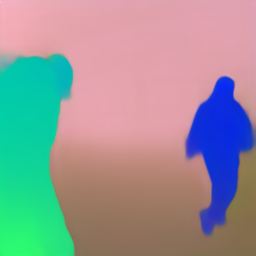} \\ [1pt]
         \includegraphics[width=\iw]{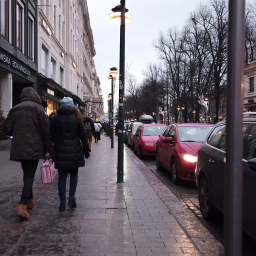} &
         \includegraphics[width=\iw]{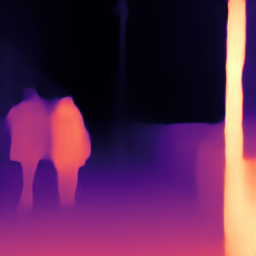} &
         \includegraphics[width=\iw]{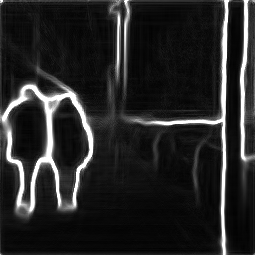} &
         \includegraphics[width=\iw]{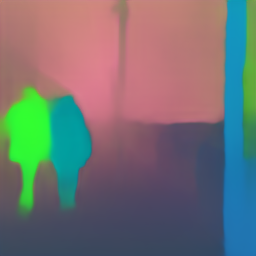} \\ [1pt]
         \includegraphics[width=\iw]{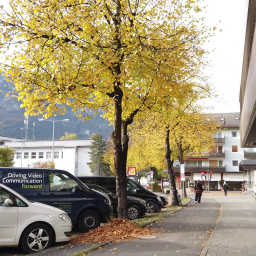} &
         \includegraphics[width=\iw]{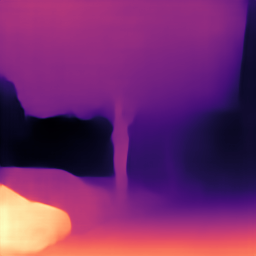} &
         \includegraphics[width=\iw]{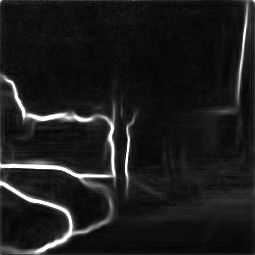} &
         \includegraphics[width=\iw]{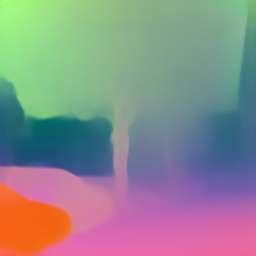} \\ [1pt]
         \includegraphics[width=\iw]{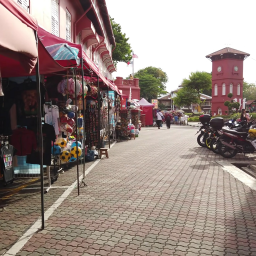} &
         \includegraphics[width=\iw]{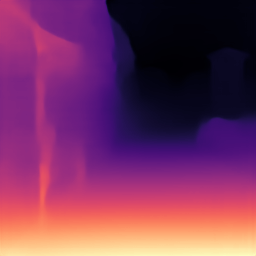} &
         \includegraphics[width=\iw]{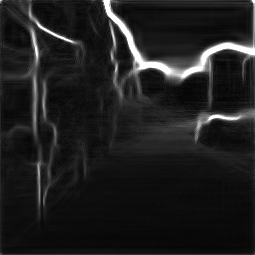} &
         \includegraphics[width=\iw]{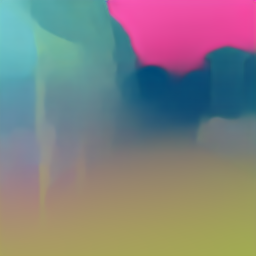} \\ [8pt]
         \scriptsize Input image & \scriptsize Predicted & \scriptsize Embedding & \scriptsize Embedding PCA \\ [4pt]
         & \scriptsize disparity & \scriptsize gradient & \scriptsize first 3 dims
    \end{tabular}
    \vspace{.5em}
    \caption{Qualitative results on Walking-Tours with visualizations of the object embedding $\embedding$. The spatial gradient of $\embedding$ (third column) separates cars and people, while also oversegmenting somewhat by drawing edges at strong depth discontinuities. The fourth column shows the first three dimensions of $\embedding$ after PCA. Instances are separated, though there is some undesirable smooth variation within individual objects and the background. See \citesec{sec:embed}. }
    \vspace{0em}
    \label{fig:walking}
\end{figure}

\para{Walking-Tours dataset.} To demonstrate the object embedding technique on scenes with moving camera and moving objects, we collect and train on a dataset of internet videos. These videos are chosen by searching for ``walking around (city)'' for each of the largest 50 cities in the world by population, and collected and processed following the same approach as the RealEstate10K dataset~\cite{Zhou:TOG18}. In total, we consider about 250 videos and about 1.2M frames. The videos, which are mostly tours shot from hand-held or vehicle-mounted cameras, are dynamic and feature both camera and object motion. They vary in geographic location, image quality, camera hardware, and resolution.

\para{Qualitative analysis.} Qualitative results are shown in \citefig{fig:walking}. We find that the network simultaneously learns disparity and a (soft) instance segmentation. Since the instance embedding vector is a unit vector in $\mathbb{R}^6$, we visualize it in three ways. In \citefig{fig:walking}, we reduce the dimension using principal component analysis~\cite{Pearson:PCA:01} and show the top three principal dimensions in RGB; we also show the (spatial) gradient magnitude of the embedding, which highlights strong edges. For example, the first two rows of~\citefig{fig:walking} show that the network has assigned different instances of people different vectors. In~\citefig{fig:walking2}, we show a proof of concept of using the embedding for instance segmentation: we (manually) choose a few seed points on each of several objects. Then, every other pixel is assigned to the closest seed in bilateral space, i.e., where the distance is the the (weighted) sum of distances in euclidean and embedding space.

\begin{figure}[t]
\centering
\renewcommand{\tabcolsep}{.5pt}
\renewcommand{\arraystretch}{0}
\newcommand{\iw}{\dimexpr 0.25\linewidth - .5pt\relax}
    \begin{tabular}{cccc}
         \includegraphics[width=\iw]{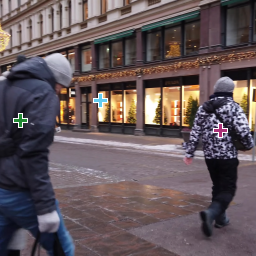} &
         \includegraphics[width=\iw]{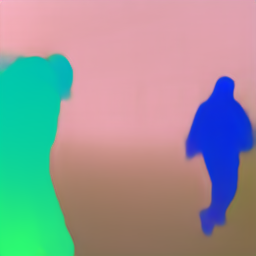} &
         \includegraphics[width=\iw]{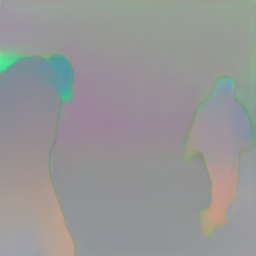} &
         \includegraphics[width=\iw]{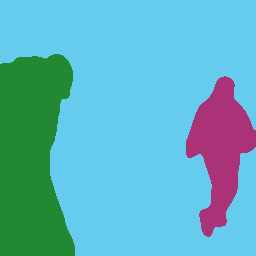} \\ [1pt]
         \includegraphics[width=\iw]{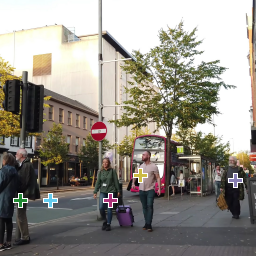} &
         \includegraphics[width=\iw]{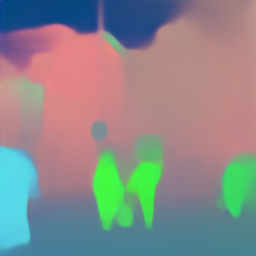} &
         \includegraphics[width=\iw]{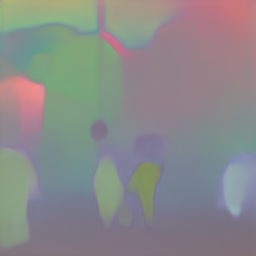} &
         \includegraphics[width=\iw]{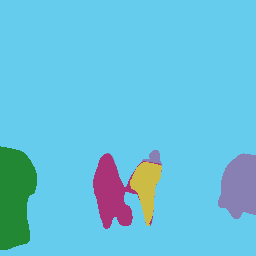} \\ [1pt]
         \includegraphics[width=\iw]{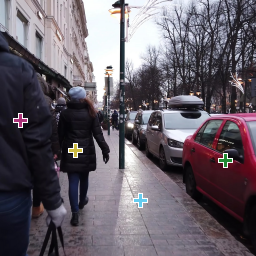} &
         \includegraphics[width=\iw]{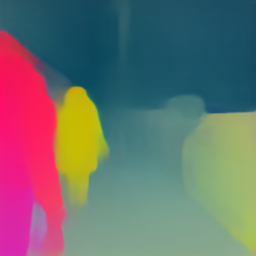} &
         \includegraphics[width=\iw]{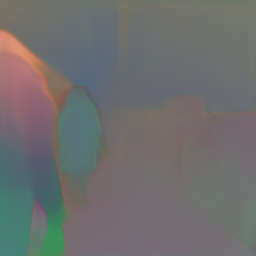} &
         \includegraphics[width=\iw]{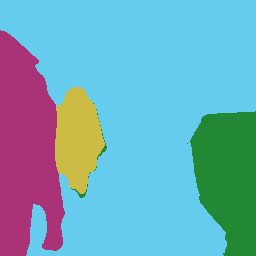} \\ [1pt]
         \includegraphics[width=\iw]{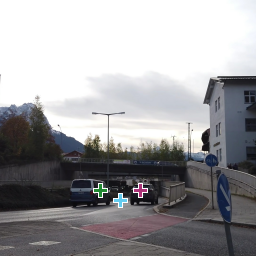} &
         \includegraphics[width=\iw]{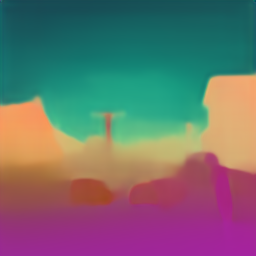} &
         \includegraphics[width=\iw]{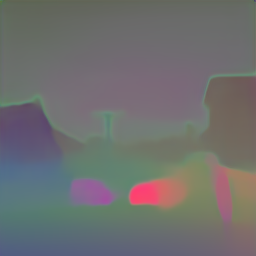} &
         \includegraphics[width=\iw]{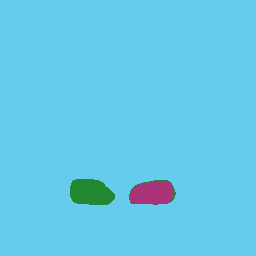} \\ [1pt]
         \includegraphics[width=\iw]{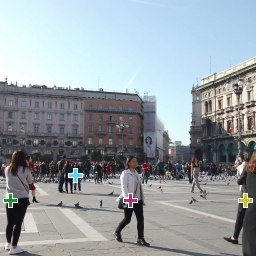} &
         \includegraphics[width=\iw]{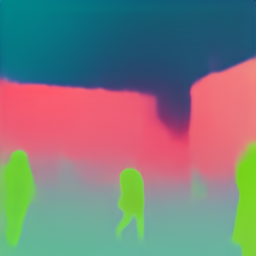} &
         \includegraphics[width=\iw]{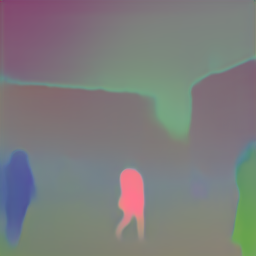} &
         \includegraphics[width=\iw]{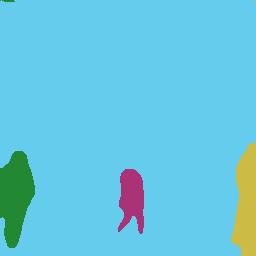} \\ [8pt]
         \scriptsize Input & \scriptsize Embedding PCA & \scriptsize Embedding PCA & \scriptsize Induced \\ [4pt]
         \scriptsize with seed points & \scriptsize (dims 1--3) & \scriptsize (dims 4--6) & \scriptsize segmentation
    \end{tabular}
    \vspace{.5em}
    \caption{Example segmentation from our instance embedding.
    We manually choose seed points for objects and background, and induce a segmentation by coloring each pixel according to which of the seed points is closest to it in bilateral embedding space.
    Using only self-supervision, the network has learned to separate person and car instances. See \citesec{sec:embed}.}
    \vspace{0em}
    \label{fig:walking2}
\end{figure}

\para{Oversegmentation.} One issue is that, beyond the limits imposed by the dimensionality of $\embedding$, the network has no incentive not to oversegment. In particular, it places embedding edges between regions where the difference between flows is hard to predict from a single image. These include object boundaries, as desired, but also includes large depth discontinuities, such as between a foreground and midground objects. Thus the network will sometimes place object-embedding edges at, e.g., the horizon, as seen in \citefig{fig:walking}.

\section{Discussion}
We show that our subspace model can be applied to the tasks of disparity estimation and object instance embedding from in-the-wild internet videos without the use of camera intrinsics/pose or multiview stereo.

Our approach has some limitations. One is that we rely on the instantaneous-flow assumption, and so our method is suitable for use only on video datasets in which the motion is not too fast. It is unlikely to be suitable for image-colletion datasets such as Megadepth~\cite{Li:megadepth:CVPR18}. Relatedly, our method depends on flow, which could be degraded by large motions, occlusion, specularities, and so on.

Our method is also limited by an independence assumption for dynamic scene content. For example, for a given image pair an object that has the same motion as the camera will have the same flow as an object at infinite depth. This situation is common in driving datasets; we find that when trained on datasets such as KITTI~\cite{Geiger:KITTI13} or Waymo~\cite{sun2020scalability}, the network tends to do well on static portions of the scene but assigns a very large depth to cars moving in the same direction as the capturing car.
Our method is more suitable to in-the-wild internet-collected datasets such as RealEstate10K or Walking-Tours, which tend to feature more general camera motions.

\clearpage
\newpage
{
\small
\bibliographystyle{ieee_fullname}
\bibliography{dimmot}
}

\clearpage
\newpage
\setcounter{section}{0}
\renewcommand*{\thesection}{\Alph{section}}
\def\down{$\Downarrow$}
\def\up{$\Uparrow$}
\newcommand\conv[1]{$\textrm{conv}_{#1}$}

\section{Network architecture}

\noindent Our network (modeled on those of [27] and [46]), is detailed in the following table:

\begin{table}[h!]
\newcommand{\ind}{\hspace{.5em}}
\centering
\begin{tabular}{l|rrrrll}
\toprule
Input & $\mathrm{k}_1$ & $\mathrm{c}_1$ & $\mathrm{k}_2$ & $\mathrm{c}_2$ & Output
\tabularnewline
\midrule
$\textrm{Norm}(\Image)$                   & 7 &  32 & 7 &  32 & \conv{1}   \tabularnewline
\ind\down(\conv{1})            & 5 &  64 & 5 &  64 & \conv{2}   \tabularnewline
\ind\ind\down(\conv{2})            & 3 & 128 & 3 & 128 & \conv{3}   \tabularnewline
\ind\ind\ind\down(\conv{3})            & 3 & 256 & 3 & 256 & \conv{4}   \tabularnewline
\ind\ind\ind\ind\down(\conv{4})            & 3 & 512 & 3 & 512 & \conv{5}   \tabularnewline
\ind\ind\ind\ind\ind\down(\conv{5})            & 3 & 512 & 3 & 512 & \conv{6}   \tabularnewline
\ind\ind\ind\ind\ind\ind\down(\conv{6})            & 3 & 512 & 3 & 512 & \conv{7}   \tabularnewline
\ind\ind\ind\ind\ind\ind\ind\down(\conv{7})            & 3 & 512 & 3 & 512 & \conv{8}   \tabularnewline

\ind\ind\ind\ind\ind\ind\up(\conv{8}) + \conv{7}\quad   & 3 & 512 & 3 & 512 & \conv{9}   \tabularnewline
\ind\ind\ind\ind\ind\up(\conv{9}) + \conv{6}   & 3 & 512 & 3 & 512 & \conv{10}  \tabularnewline
\ind\ind\ind\ind\up(\conv{10}) + \conv{5}  & 3 & 512 & 3 & 512 & \conv{11}  \tabularnewline
\ind\ind\ind\up(\conv{11}) + \conv{4}  & 3 & 256 & 3 & 256 & \conv{12}  \tabularnewline
\ind\ind\up(\conv{12}) + \conv{3}  & 3 & 128 & 3 & 128 & \conv{13}  \tabularnewline
\ind\up(\conv{13}) + \conv{2}  & 3 &  64 & 3 &  64 & \conv{14}  \tabularnewline
\up(\conv{14}) + \conv{1}  & 3 &  64 & 3 &  64 & \conv{15}  \tabularnewline
\conv{15}                  & 3 &  32 & 3 &  32 & \conv{16}  \tabularnewline
\conv{16}                  & 3 &  $C$ & - & - & output  \tabularnewline
\bottomrule
\end{tabular}
\end{table}
Each row above (except the last) describes a pair of convolutional layers in sequence with kernel sizes $\mathbf{k}_1, \mathbf{k}_2$ and number of output channels $\mathbf{c}_1, \mathbf{c}_2$. \textbf{Input} shows the input to the first layer, where $\textrm{Norm}$ denotes ImageNet-style normalization, \down\ denotes maxpooling with a pool size of 2 (thus halving the size), \up\ denotes nearest-neighbour upscaling by a factor of 2, and $+$ is concatenation. Each layer is followed by ReLU activation.

The final row shows a single convolutional layer which outputs $C$ channels. In our disparity experiments, $C = 1$ and is followed by sigmoid activation. In our disparity plus embedding experiments, $C = 7$: one channel for disparity (with sigmoid activation) and six for embedding (normalized to be unit-length at each pixel).

\section{SVD details}

To compute $\flowhat$, as described in Section~3, we assemble the matrix whose column space is $\Space$:
\begin{equation}
    M = \begin{bmatrix} \flow_0 \vert \flow_1 \vert \cdots \vert \flow_{n-1} \end{bmatrix},
\end{equation}
the dimensions of which are $2HW \times n$. Before assembling $M$, we normalize rotational basis vectors to have norm 1 and translational basis vectors to have norm 2 (prior to pointwise multiplication by disparity). We compute the singular-value decomposition of $M$:
\begin{equation}
    M = U\Sigma V^T.
\end{equation}
We choose the columns of $U$ corresponding to singular values (entries in $\Sigma$) greater than a threshold $\varepsilon$ (in our experiments, $\varepsilon=\num{1e-5}$); calling this submatrix $U_{s}$ we compute $\flowhat$ via
\begin{equation}
    \flowhat = U_sU_s^T \flow.
\end{equation}

\section{Training details}

We use the ADAM optimizer with a learning rate of \num{5e-5} and an L2 regularization on network weights of \num{1e-6}, and train asynchronously using ten workers with a batch size of 4 per worker. In those experiments which learn an object embedding, we project the ground truth flow twice and compute two losses: one using the basis only resulting from the learned disparity (loss weight $0.5$), and one using the full projection described in Section~3.4 (loss weight $1.0$). In experiments without object embedding, flow reconstruction loss has a weight of $1.0$.

We found in training that the network sometimes produces very large values for disparity or instance embedding; we apply a regularization loss on disparity before sigmoid activation of $L(z) = \text{max}(0, z-5)$; and a loss on instance-embedding before normalization, $L(z) = \text{max}(0, (\sum_i z_i^2)-1)$. Each of these are averaged over the image and applied with a weight of \num{1e-6}. We train for about 5M steps. We choose the best model and checkpoint from five runs based on flow reprojection loss on a held-out validation set.

\end{document}